\documentclass{llncs}

\newcommand{\putdisk}[3]{\put(#1,#2){\circle*{#3}}}
\newcommand{\puttext}[3]{\put(#1,#2){{\mbox{\scriptsize$#3$\normalsize}}}}

\newcommand{\type}{ \,{\bf :}\, }
\newcommand{\product}[2]{ {#1} {\times} {#2} }

\newcommand{\pair}[2]{\langle #1,#2 \rangle}

\newcommand{\triple}[3]{\mbox{$ \langle #1,#2,#3 \rangle $}}
\newcommand{\quadruple}[4]{\mbox{$ \langle #1,#2,#3,#4 \rangle $}}

\begin{document}

\title{Conceptual Analysis of Hypertext}

\author{Robert E.\ Kent\inst{1}\thanks{This research was funded 
by a grant from Intel Corporation.}
and Christian Neuss\inst{2}}

\institute{
Washington State University,
Pullman, WA 99164, USA
\and
Technische Hochschule Darmstadt,
64289 Darmstadt,
Germany}

\maketitle

\begin{abstract}
In this chapter 
tools and techniques 
from the mathematical theory of formal concept analysis 
are applied to hypertext systems in general, 
and the World Wide Web in particular. 
Various processes for the conceptual structuring of hypertext are discussed: summarization,
conceptual scaling, 
and the creation of conceptual links. 
Well-known interchange formats 
for summarizing networked information resources 
as resource meta-information are reviewed, 
and two new interchange formats 
originating from formal concept analysis are advocated. 
Also reviewed is conceptual scaling, 
which provides a principled approach 
to the faceted analysis techniques in library science classification. 
The important notion of conceptual linkage 
is introduced as a generalization of a hyperlink. 
The automatic hyperization of the content of legacy data is described, 
and the composite conceptual structuring with hypertext linkage is defined.
For the conceptual empowerment of the Web user,
a new technique called conceptual browsing is advocated. 
Conceptual browsing, 
which browses over conceptual links, 
is dual mode (extensional versus intensional) 
and dual scope (global versus local).
\end{abstract}


\section{Conceptual Knowledge Systems}

Using ideas from library science \cite{Ro87,WyTa92},
hypertext systems \cite{Wi94},
and formal concept analysis \cite{Wi82,GaWi89}, 
tools are currently being developed \cite{KeNe94,NeKe95,KeBo95}
for the conceptual analysis of networked information resources in general, and the World Wide Web in particular. 
Networked information resources include 
(1) individual text files, 
(2) WAIS databases, and 
(3) starting points for hypertext webs. 

Resources are best thought of, not as objects, 
but as conceptual classes (formal concepts). 
We offer a concept-oriented approach 
for the description and organization of networked information resources, which will facilitate their subsequent discovery and access. 
This should not be thought of as yet another object-oriented approach. Although objects generate their own classes, 
classes are not only more general but also include intensional information. By identifying concepts with classes, 
this can be regarded as a class-oriented approach 
--- an approach that has been advocated recently by Terry Winograd 
in the IETF-URI working group discussion on library standards and URI, 
and supported by Ronald Daniel and Dirk Herr-Hoyman.

Formal concept analysis \cite{Wi82,GaWi89} is a relatively new discipline 
arising out of 
the mathematical theory of lattices 
and 
the calculus of binary relations. 
It is closely related to the areas of knowledge representation 
in computer science and cognitive psychology. 
Formal concept analysis provides 
for the automatic classification of both knowledge and documents 
via representation of a user's faculty for interpretation 
as encoded in conceptual scales. 
Such conceptual scales correspond to 
the facets of synthetic classification schemes, 
such as Ranganathan's Colon classification scheme, in library science.

Formal concept analysis uses objects, attributes and conceptual classes
as its basic constituents.
Objects and attributes are connected through 
has-a incidence relationships,
while
conceptual classes are connected through 
is-a subtype relationships.
Incidence is the most primitive notion in formal concept analysis.
A {\em formal context\/} represents incidence by 
collecting together all of the relevant has-a relationships.
It is a triple $\triple{G}{M}{I}$
consisting of 
a set of objects $G$ (Gegenst\"{a}nde, in German),
a set of attributes $M$ (Merkmale, in German),
and a binary incidence relation $I \subseteq \product{G}{M}$,
where $g{I}m$ asserts that ``object $g$ has attribute $m$.''
In many contexts appropriate for Web resources,
the objects are document-like objects
and 
the attributes are properties of those document-like objects 
which are of interest to the Web user.

A {\em conceptual class\/} or {\em formal concept\/}
is the central notion in formal concept analysis.
A formal concept consists of 
a collection of entities or objects
exhibiting one or more common characteristics, traits or attributes.
Conceptual classes are logically characterized by 
their extension and intension.
The {\em extension\/} of a class is 
the aggregate of entities or objects
which it includes or denotes.
The {\em intension\/} of a class is 
the sum of its unique characteristics, traits or attributes,
which, taken together,
imply the concept signified by the conceptual class.
In this paper
conceptual classes are identified with the concept which they signify.
The process of subordination of concepts and collocation of objects
exhibits a natural order,
proceeding top-down
from 
the more generalized concepts with larger extension and smaller intension
to 
the more specialized concepts with smaller extension and larger intension.
This is-a relationship is a partial order called generalization-specialization.
Concepts with this generalization-specialization ordering
form a class hierarchy ${\cal L} = {\cal L}\triple{G}{M}{I}$
called a {\em concept lattice\/}.
Formal concept analysis 
uses formal concepts as its central notion
and 
uses concept lattices as an approach to knowledge representation \cite{Wi82}.
The use of conceptual classes as a conceptual structuring mechanism
corresponds to
the use of similarity clusters in information retrieval \cite{Wi94},
although conceptual classes are based more on logical implication
rather than a nearness notion.
However,
see the discussion about conceptual linkage below.

The enriching notion of a {\em conceptual knowledge system\/} 
from formal concept analysis \cite{Wi92,KeBo95}
allows,
not only the modeling of knowledge representation,
but also the ability to do 
knowledge inferencing, knowledge acquisition, and knowledge communication. 
In a conceptual knowledge system
there are three basic notions: 
objects, attributes, and conceptual views. 
These are connected through four basic relationships: 
an object has an attribute (incidence), 
an object belongs to a conceptual view (instantiation), 
an attribute abstracts from a conceptual view (abstraction), 
and 
a conceptual view is a subordinate to another conceptual view (subtype).
These notions and relationships partition
the frame of a conceptual knowledge system
as in Table~\ref{fca:relationships}.
In a conceptual knowledge system 
we distinguish between
(1) anonymous concepts
which are automatically and implicitly generated 
from the four basic relationships
and represent a form of conceptual resource discovery,
and
(2) named and explicitly specified concepts
which we call conceptual views.
Compare the distinct, but closely related, notion 
of a Nebula-style view \cite{KeBo95}.

\begin{table}
\caption{Conceptual Knowledge System Relationhips}
\label{fca:relationships}
\begin{center}
\fbox{\begin{tabular}{|r||c|c|}\hline
			& Views		& Attributes	\\ \hline\hline
	Views		& subtype	& abstraction	\\ \hline
	Objects		& instantiation	& incidence	\\ \hline
\end{tabular}}
\end{center}
\end{table}

Table~\ref{document:cks} represents a conceptual knowledge system 
within the conceptual universe ${\cal D}$ 
of all documents in an information system and their properties
(see Figure 2 in \cite{BDBCP94}).
In addition to a set of document-like objects and attributes,
it contains the five conceptual views
$\{ 
{\rm Object}, {\rm Document}, {\rm PostScript}, {\rm Plan1}, {\rm Plan2}
\}$.
The crosses in the table of basic relationships in Table~\ref{document:cks}
are partitioned
into the four parts described in Table~\ref{fca:relationships}: 
subtype, in the upper left;
abstraction, in the upper right;
instantiation, in the lower left; and
incidence, in the lower right.
The bottom panel of Table~\ref{document:cks} is 
the line diagram of the lattice of conceptual classes,
which represents the conceptual space for the document
conceptual knowledge system.

The representational mechanism of conceptual knowledge systems
serves as a firm foundation for 
the basic paradigms of
internet resource discovery and wide area information management systems:
organization-navigation and search-retrieval \cite{KeBo95}.
The use of conceptual knowledge systems is a natural outgrowth
of the original formal concept analysis approach
for structuring and organizing the networked information resources
in the World Wide Web \cite{KeNe94}.

\footnotesize
\begin{table}
\caption{Conceptual Knowledge System in the Document Universe}
\label{document:cks}
\begin{center}
\begin{tabular}{c}
\begin{tabular}{c@{\hspace{5mm}}c@{\hspace{5mm}}c}
\begin{tabular}[t]{|r@{\hspace{2mm}}l|} \hline
	\multicolumn{2}{|c|}{Views/Objects} \\ \hline\hline
	1 & Object     \\
	2 & Document   \\
	3 & PostScript \\
	4 & Plan1      \\
	5 & Plan2      \\ \hline
	6 & plan1.ps   \\
	7 & plan2.ps   \\
	8 & plan2.doc  \\
	9 & notes0.txt \\
	10 & notes1.txt \\
	11 & notes2.txt \\ \hline
\end{tabular}
&
\begin{tabular}[t]{|r|c@{\hspace{3pt}}c@{\hspace{3pt}}c@{\hspace{3pt}}c@{\hspace{3pt}}
	c@{\hspace{3pt}}|c@{\hspace{3pt}}c@{\hspace{3pt}}c@{\hspace{3pt}}c|} \hline
	\multicolumn{10}{|c|}{Basic Relationships} \\ \hline\hline
	           & 1 & 2 & 3 & 4 & 5 & 6 & 7 & 8 & 9 \\ \hline
	1 &$\times$&&&& &&&& \\
	2 &$\times$&$\times$&&& &&&& \\
	3 &$\times$&$\times$&$\times$&& &&&$\times$& \\
	4 &$\times$&$\times$&&$\times$& &$\times$&&& \\
	5 &$\times$&$\times$&&&$\times$ &&$\times$&& \\ \hline
	6 &$\times$&$\times$&$\times$&$\times$& &$\times$&&$\times$& \\
	7 &$\times$&$\times$&$\times$&&$\times$ &&$\times$&$\times$& \\
	8 &$\times$&$\times$&&&$\times$ &&$\times$&& \\
	9  &$\times$&$\times$&&$\times$&&$\times$&&&$\times$ \\
	10 &$\times$&$\times$&&&$\times$&&$\times$&&$\times$ \\
	11 &$\times$&$\times$&&&$\times$&&$\times$&&$\times$ \\ \hline
\end{tabular}
&
\begin{tabular}[t]{|r@{\hspace{2mm}}l|} \hline
	\multicolumn{2}{|c|}{Views/Attributes} \\ \hline\hline
	1 & Object            \\
	2 & Document          \\
	3 & PostScript        \\
	4 & Plan1             \\
	5 & Plan2             \\ \hline
	6 & project=plan1     \\
	7 & project=plan2     \\
	8 & format=postscript \\
	9 & format=text       \\ \hline
\end{tabular}
\end{tabular}
\\ \\ \\
\setlength{\unitlength}{1pt}
\begin{tabular}{|c|}\hline
Concept Lattice \\ \hline\hline
\begin{picture}(290,205)
\put(45,15){\begin{picture}(200,175)
	\puttext{0}{0}{{\bf }}
	\putdisk{100}{175}{8}				
	\puttext{105}{171}{\framebox(25,10){\rm Object}}
	\putdisk{100}{150}{8}				
	\puttext{106}{146}{\framebox(38,10){\rm Document}}
	\put(100,150){\line(0,1){25}}			
	\putdisk{25}{100}{8}				
	\puttext{0}{106}{{\rm format{=}postscript}}	
	\puttext{31}{93}{\framebox(38,10){\rm PostScript}}
	\put(25,100){\line(3,2){75}}			
	\putdisk{75}{100}{8}				
	\puttext{70}{106}{{\rm project{=}plan2}}	
	\puttext{81}{93}{\framebox(25,10){\rm Plan2}}	
	\puttext{80}{86}{{\rm plan2.doc}}		
	\put(75,100){\line(1,2){25}}			
	\putdisk{125}{100}{8}				
	\puttext{125}{106}{{\rm project{=}plan1}}	
	\puttext{131}{93}{\framebox(25,10){\rm Plan1}}	
	\put(125,100){\line(-1,2){25}}			
	\putdisk{175}{100}{6}				
	\puttext{179}{105}{{\rm format{=}text}}		
	\put(175,100){\line(-3,2){75}}			
	\putdisk{0}{50}{6}				
	\puttext{5}{44}{{\rm plan2.ps}}			
	\put(0,50){\line(1,2){25}}			
	\put(0,50){\line(3,2){75}}			
	\putdisk{50}{50}{6}				
	\puttext{55}{44}{{\rm plan1.ps}}		
	\put(50,50){\line(-1,2){25}}			
	\put(50,50){\line(3,2){75}}			
	\putdisk{150}{50}{6}				
	\puttext{155}{46}{{\rm notes1.txt}}		
	\puttext{155}{39}{{\rm notes2.txt}}		
	\put(150,50){\line(-3,2){75}}			
	\put(150,50){\line(1,2){25}}			
	\putdisk{200}{50}{6}				
	\puttext{205}{44}{{\rm notes0.txt}}		
	\put(200,50){\line(-3,2){75}}			
	\put(200,50){\line(-1,2){25}}			
	\putdisk{100}{0}{6}				
	\put(100,0){\line(-2,1){100}}			
	\put(100,0){\line(-1,1){50}}			
	\put(100,0){\line(1,1){50}}			
	\put(100,0){\line(2,1){100}}			
\end{picture}}
\end{picture}
\\ \hline
\end{tabular}
\end{tabular}
\end{center}
\end{table}
\normalsize


\section{Resource Meta-information}

Due to the rapid growth of the World Wide Web,
resource discovery has become a serious problem.
Because of its decentralized architecture,
the user experiences the Web as a large information repository 
without an underlying structure.
The process of ``surfing'' pages by repeatedly following hyperlinks
is the most popular use of the Web.
It can however lead to the phenomenon of
getting ``lost in hyperspace.''

From the very beginning, 
approaches have been made to organize information
about networked information resources into catalogs and indexes. 
Index files were originally maintained manually. 
However, 
the rapid growth of the Web soon made necessary 
automatic methods for generating resource directories. 
Automatic tools called ``robots'', ``Web wanderers'' or ``spiders'' 
soon evolved. 
These are programs 
which automatically 
connect to a remote server and recursively retrieve documents. 
Since Web robots often put heavy loads on Web servers, 
they have been controversial, 
and are sometimes disliked by server maintainers.

Web robots are trailing-edge technologies.
The main problem with robots is that 
they are not true Web wanderers 
--- the retrieval program does not transfer itself 
from the index site to the provider site, 
but instead it transfers in the reverse direction over the network 
all the potentially indexable documents. 
Since document repositories may contain hundreds of megabytes, 
the bandwidth requirements are enormous. 
Exacerbating this problem is the fact that 
current indexing tools gather independently, 
without sharing information with other indexers.

A partial answer to these problems are
Networked Information Discovery and Retrieval (NIDR) systems
such as Harvest \cite{BDHMS94}.
A more complete answer will involve
NIDR systems with conceptual structuring mechanisms\cite{KeNe94}
such as the WAVE\footnote{The first author is the principal investigator for an Intel funded project
which is developing and assessing the WAVE system.}
system 
(Web Analysis and Visualization Environment)
which is being developed by following principles espoused in this chapter.
NIDR systems are leading-edge technologies
which 
reduce the load on information servers,
reduce network traffic,
and 
reduce index disk space requirements,
principally by use of resource meta-information
(also called metadata)
--- they extract meta-information at the provider site,
sending this, and not the raw data, over the network.
This section reviews various formats 
used by NIDR systems and library science
for representing resource meta-information
as bibliographic records \cite{NeKe95}.

\begin{description}
\sloppy
\item[Uniform Resource Characteristics:]
The on-going discussions concerning metadata
in various internet engineering task force (IETF) working groups
are centered around the following notions \cite{Me94}.
A Uniform Resource Locator (URL) 
is used for hyperlink markup in Web documents. 
Since a URL specifies 
a location and retrieval protocol of a given networked information resource, 
it is not a long-lived, stable reference.
A Uniform Resource Name (URN) is used to identify a resource. 
It is long-lived and persistent,
and uniquely names a networked information resource. 
A Uniform Resource Locator 
is used to locate an instance of a resource identified by an URN. 
A Uniform Resource Characteristic (URC) 
is used to represent URNs with associated meta-information. 
URCs are analogous to the bibliographic records of library science. 
URCs encode meta-information about network resources in the form of attribute-value pairs.
\item[IAFA Templates:]
The internet anonymous ftp archives (IAFA) working group of the IETF 
has proposed a format for indexing information 
that can be used to describe various internet resources. 
The IAFA template specification \cite{DeEm94}
encodes pieces of meta-information.
The IAFA templates are intended to be both human and machine readable.
Archie servers support this format to provide information about items
available for anonymous ftp. 
Work is currently underway for the
construction of Uniform Resource Identifiers.
\item[Harvest Summary Object Interchange Format:]
Harvest is a set of tools 
to gather, extract, and search relevant information 
across the internet \cite{BDHMS94}. 
It provides methods for
distributed indexing, building topic specific indices, flexible search
strategies, and replica systems. 
Harvest generates a content summary for each information object it gathers.
These records are stored in a format 
called the Summary Object Interchange Format (SOIF).
SOIF is based on a combination of the IAFA templates and {\sc Bib}\TeX.
\item[Bibliographic Records from Library Science:]
In order to compare URCs, IAFA templates, and Harvest SOIFs
with bibliographic description in library science,
listed here are some attributes,
which are classified according to the eight areas 
of the international standards for bibliographic description
(ISBD) \cite{WyTa92}:
title and statement of responsibility (title, author);
edition (version);
material (or type of publication) specific details;
publication, distribution, etc.;
physical description (content-type, content-length, size, cost, etc.);
series (time-to-live);
notes (abstract); 
and
standard number and terms of availability 
(uniform resource name, uniform resource locator).
\fussy
\end{description}

Table~\ref{formats} lists two generic interchange formats
which can be used to specify
faceted information in conceptually scaled 
networked information resources \cite{NeKe95}.
Such faceted information can occur
in various interfaces in a resource discovery system. 
From a mathematical viewpoint,
these two representations are equivalent to each other. 
Software exists for converting between the two forms. 

The left side of Table~\ref{formats}
displays the Formal Context Interchange Format (FCIF).
FCIF is oriented towards the formal contexts of formal concept analysis.
FCIF represents order-theoretic 
formal contexts of networked information resources, 
consisting of two partially ordered sets, 
a poset of objects and a poset of single-valued attributes, 
and an order-preserving incidence matrix 
which represents the has relationship between objects and attributes. 
The right side of Table~\ref{formats} 
displays the Concept Lattice Interchange Format (CLIF). 
CLIF is oriented towards 
the concept lattices of formal concept analysis.
CLIF provides a storage-optimal representation 
of order-theoretic lattices of conceptual classes 
for networked information resource meta-information, 
consisting of (the inverse relationships for)
two generator monotonic functions, 
from the posets of objects and attributes 
to the lattice of conceptual classes, 
and a successor matrix
which represents the subtype relationship between conceptual classes.

\small
\begin{table}
\caption{Interchange Formats for Faceted Resource Meta-information}
\label{formats}
\begin{center}
\begin{tabular}[t]{c}
\begin{tabular}[t]{c@{\hspace{5mm}}c}
\begin{tabular}[t]{|c|}\hline
	Formal Context Interchange Format \\ \hline\hline \\
\begin{tabular}{l}
{\tt TYPE} \\
\mbox{\hspace{6mm}} $T$ \\
{\tt OBJECT} \\
\mbox{\hspace{6mm}} $O_1$ \verb|{| $O_{1,1}\; O_{1,2}\; \cdots\; O_{1,o_1}$ \verb|}| \\
\mbox{\hspace{6mm}} $O_2$ \verb|{| $O_{2,1}\; O_{2,2}\; \cdots\; O_{2,o_2}$ \verb|}| \\
\mbox{\hspace{6mm}} $\cdots$ \\
\mbox{\hspace{6mm}} $O_n$ \verb|{| $O_{n,1}\; O_{n,2}\; \cdots\; O_{n,o_n}$ \verb|}| \\
{\tt ATTRIBUTE} \\
\mbox{\hspace{6mm}} $A_1$ \verb|{| $A_{1,1}\; A_{1,2}\; \cdots\; A_{1,a_1}$ \verb|}| \\
\mbox{\hspace{6mm}} $A_2$ \verb|{| $A_{2,1}\; A_{2,2}\; \cdots\; A_{2,a_2}$ \verb|}| \\
\mbox{\hspace{6mm}} $\cdots$ \\
\mbox{\hspace{6mm}} $A_m$ \verb|{| $A_{m,1}\; A_{m,2}\; \cdots\; A_{m,a_m}$ \verb|}| \\
{\tt INCIDENCE} \\
\mbox{\hspace{6mm}} $O_1$ \verb|{| $A_{1,1}\; A_{1,2}\; \cdots\; A_{1,i_1}$ \verb|}| \\
\mbox{\hspace{6mm}} $O_2$ \verb|{| $A_{2,1}\; A_{2,2}\; \cdots\; A_{2,i_2}$ \verb|}| \\
\mbox{\hspace{6mm}} $\cdots$ \\
\mbox{\hspace{6mm}} $O_n$ \verb|{| $A_{n,1}\; A_{n,2}\; \cdots\; A_{n,i_n}$ \verb|}| \\
\end{tabular}
\\ \\ \hline
\end{tabular}
&
\begin{tabular}[t]{|c|}\hline
	Concept Lattice Interchange Format \\ \hline\hline \\
\begin{tabular}{l}
{\tt TYPE} \\
\mbox{\hspace{6mm}} $T$ \\
{\tt GENERATOR:OBJECT} \\
\mbox{\hspace{6mm}} $C_1$ \verb|{| $O_{1,1}\; O_{1,2}\; \cdots\; O_{1,o_1}$ \verb|}| \\
\mbox{\hspace{6mm}} $C_2$ \verb|{| $O_{2,1}\; O_{2,2}\; \cdots\; O_{2,o_2}$ \verb|}| \\
\mbox{\hspace{6mm}} $\cdots$ \\
\mbox{\hspace{6mm}} $C_p$ \verb|{| $O_{p,1}\; O_{p,2}\; \cdots\; O_{p,o_p}$ \verb|}| \\
{\tt GENERATOR:ATTRIBUTE} \\
\mbox{\hspace{6mm}} $C_1$ \verb|{| $A_{1,1}\; A_{1,2}\; \cdots\; A_{1,a_1}$ \verb|}| \\
\mbox{\hspace{6mm}} $C_2$ \verb|{| $A_{2,1}\; A_{2,2}\; \cdots\; A_{2,a_2}$ \verb|}| \\
\mbox{\hspace{6mm}} $\cdots$ \\
\mbox{\hspace{6mm}} $C_p$ \verb|{| $A_{p,1}\; A_{p,2}\; \cdots\; A_{p,a_p}$ \verb|}| \\
{\tt SUCCESSOR} \\
\mbox{\hspace{6mm}} $C_1$ \verb|{| $C_{1,1}\; C_{1,2}\; \cdots\; C_{1,s_1}$ \verb|}| \\
\mbox{\hspace{6mm}} $C_2$ \verb|{| $C_{2,1}\; C_{2,2}\; \cdots\; C_{2,s_2}$ \verb|}| \\
\mbox{\hspace{6mm}} $\cdots$ \\
\mbox{\hspace{6mm}} $C_p$ \verb|{| $C_{p,1}\; C_{p,2}\; \cdots\; C_{p,s_p}$ \verb|}| \\
\end{tabular}
\\ \\ \hline
\end{tabular}
\end{tabular}
\\ \\
\begin{tabular}{c} \\
\begin{minipage}{10.7cm}
\begin{itemize}
	\item $O_i$ and $O_{i,o}$ are object names (strings).
	\item $A_i$ and $A_{j,a}$ are attributes {\it tag\/}\#{\it value\/}, where \# is $=$, $\leq$, etc.
	\item $C_k$ and $C_{k,s}$ are indexes (natural numbers) of conceptual classes.
	\item $x_i$ and $y_j$     are coordinates (natural numbers) of conceptual class nodes.
\end{itemize}
\end{minipage}\\
\end{tabular}
\end{tabular}
\end{center}
\end{table}
\normalsize

FCIF and CLIF subsume both 
the URCs of the IETF
and 
the SOIFs of Harvest. 
The FCIF and CLIF interchange formats are more general mechanisms
than either URCs or SOIFs, 
and allow for the specification of 
more complex conceptually structured systems of resources. 
Actually, as Figure 3 points out, 
both FCIF and CLIF are better thought to occur after conceptual scaling,
whereas both URC and SOIF specify ``raw meta-information'' 
which exists before conceptual scaling \cite{GaWi89}. 

From the philosophical viewpoint of formal concept analysis, 
conceptual scaling is an act of interpretation.
It maps raw uninterpreted data, 
such as occurs in URC or SOIF, 
into the end-user's conceptual scheme. 
URC and SOIF represent database entity relations, 
whereas 
FCIF represents has-a incidence relationships
between objects and attributes
and
CLIF represents is-a subtype relationships
between conceptual classes. 
These attributes are simple structured queries of the form tag\#value, 
where \# is any relational operator $=$, $\leq$, etc. 
The equality operator represents nominal scaling, 
whereas the inequality operator represents ordinal scaling \cite{GaWi89}. Through conceptual scaling, 
which often is just nominal or ordinal scaling, 
we can compare FCIF and CLIF with URC and SOIF.

\begin{figure}
\caption{Conceptual Scaling with Various Interchange Formats}
\label{interchange}
\begin{center}
\begin{tabular}{cc}
\begin{minipage}{6cm}
\begin{center}
\begin{tabular}{|c|}\hline
Conceptual Scaling with {\tt URC} \\ \hline\hline
\begin{picture}(144,40)(30,35)
		\put(48,63){{\tt FCIF}}
		\put(72,60){\vector(-1,0){30}}
		\put(48,43){{\tt CLIF}}
		\put(72,40){\vector(-1,0){30}}
	\put(72,50){\fbox{
		\begin{tabular}{c}
			Conceptual \\
			Scaling
		\end{tabular}
	}}
		\put(164,50){\vector(-1,0){30}}
		\put(144,53){{\tt URC}}
\end{picture} \\ \hline
\end{tabular}
\end{center}
\end{minipage}
&
\begin{minipage}{6cm}
\begin{center}
\begin{tabular}{|c|}\hline
Conceptual Scaling with {\tt SOIF} \\ \hline\hline
\begin{picture}(144,40)(30,35)
		\put(48,63){{\tt FCIF}}
		\put(72,60){\vector(-1,0){30}}
		\put(48,43){{\tt CLIF}}
		\put(72,40){\vector(-1,0){30}}
	\put(72,50){\fbox{
		\begin{tabular}{c}
			Conceptual \\
			Scaling
		\end{tabular}
	}}
		\put(164,50){\vector(-1,0){30}}
		\put(140,53){{\tt SOIF}}
\end{picture} \\ \hline
\end{tabular}
\end{center}
\end{minipage}
\end{tabular}
\end{center}
\end{figure}
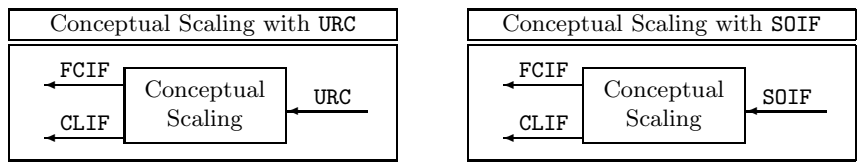


\section{Conceptual Linkage}

The structuring primitive for the World Wide Web is the hyperlink.
The essence of a hyperlink is a (possibly typed) binary association
between two objects \cite{tbl92}.
The semantics of a hyperlink is that 
the two connected objects have something in common
--- a property or a semantic category \cite{Wi94}.
By extending ideas from the field of formal concept analysis \cite{Wi82},
in this chapter we offer a principled approach 
for elevating the notion of a hyperlink
from objects to conceptual classes (concepts).
These new extended linkage structures,
which preserve the idea that
things are linked through shared attributes,
are called {\em conceptual links\/}.
The notion of conceptual links derived here
can be compare to 
similar notions in hypertext systems \cite{Wi94},
which are intuitive but not principled.
The crisp notion of a conceptual link
is represented here by
the richer, graded notion of {\em conceptual linkage\/}.
Conceptual linkage is a fuzzy relationship.
It gives a measure of similarity and implication between concepts.
Conceptual linkage can be reduced to conceptual links
by a crispification operation.
The crucial idea of conceptual linkage is derived 
from an extended theory of conceptual knowledge systems.

Conceptual linkage can be used as the structuring primitive
for the conceptual organization of the knowledge 
implicit in the World Wide Web.
There are important parallels
between conceptual knowledge systems and hypertext systems.
In particular,
conceptual links are analogous to Web hyperlinks.
Actually,
this is more than an analogy,
since objects 
(or their abstracted synoptic surrogates in the form of metadata objects) 
generate conceptual classes.
This impels us to make the following observations.
\begin{em}\begin{itemize}
	\item Networked resources are concepts (conceptual classes).
	\item Conceptual linkage extends and enriches Web hyperlinks.
	\item Conceptual space customizes and makes coherent Web hyperspace.
\end{itemize}\end{em}

There are two modes for conceptual linkage:
extensional and intensional.
Since these are dual notions in lattices,
we only discuss the extensional mode here.
In the {\em extensional mode\/} of conceptual linkage,
concepts are regarded as attributes
and are represented by their extent.
Any two concepts in extensional mode
are linked by the objects which they share,
the objects common to their extents.
The more linking objects there are,
the closer are those concepts
and
the stronger is the conceptual linkage.
This closeness can be measured by the cardinality
of the set of linking objects\footnote{For any cardinality,
we count only a kind of atomic concept called an irreducible concept:
join irreducible for object concepts
and
meet irreducible for attribute concepts.
In a concept lattice,
an object (that is, an object concept) is join irreducible
when
it cannot be decomposed as the join of two other objects,
and
an attribute (attribute concept) is meet irreducible
when
it cannot be decomposed as the meet of two other attributes.
For atomicity to be realizable,
we assume that formal concept analysis optimization
processes of purification and reduction have been carried out.
Purification fuses objects which generate the same concept.
With respect to this conceptual knowledge system,
these objects are indiscernible and equivalent.
Purification does the same for attributes.
Reduction converts objects and attributes which are not irreducible
into conceptual views.}.

The {\em extensional similarity\/} measure
$\sigma_\bullet 
 \type 
 \product{{\cal L}}{{\cal L}} \rightarrow \aleph = \{0,1,\cdots\}$
is a measure of the similarity of any two concepts 
according to their common extent cardinality.
It is the composite
	$\left( \mbox{meet} \right)
	\circ
	\left( \mbox{extent} \right)
	\circ
	\left( \mbox{cardinality} \right)$,
and is defined by the formulae
\small
\[ \sigma_\bullet(k_0,k_1) 
= \| \mbox{extent}\left({k_0 \wedge k_1}\right) \|
= \| \mbox{extent}(k_0) \cap \mbox{extent}(k_1) \| \]
\normalsize
for any two concepts $k_0,k_1$ in a concept lattice ${\cal L}$.
The bounds on this measure are
$0 
 \leq 
 \sigma_\bullet(k_0,k_1) 
 \leq 
 \mbox{min}\left\{ 
 \| \mbox{extent}(k_0) \|, \| \mbox{extent}(k_1) \| 
 \right\}$.
The extensional similarity between two concepts
is 
a rough (fuzzy?) measure of their similarity.
The closer the concepts, 
the larger the extensional similarity,
up to a maximum size of the extent cardinality of either.
When this upper bound is reached,
one concept is below the other in the concept lattice
\begin{center}$\begin{array}{r@{\hspace{4mm}}c@{\hspace{4mm}}l} 
	k_0 \leq k_1
	& \mbox{iff}
	& \| \mbox{extent}(k_0) \| =  \sigma_\bullet(k_0,k_1).
\end{array}$\end{center}
The more dissimilar the concepts,
the smaller the extensional similarity,
with lower bound 0.
When this bound is reached,
the two concepts have nothing extensionally in common.
For browsing over 
the conceptual space of a conceptual knowledge system,
we take a state space approach
where we regard concepts as conceptual states
and browsing as state transition.
Since extensional similarity is symmetric
$\sigma_\bullet(k_0,k_1) = \sigma_\bullet(k_1,k_0)$,
it does not accurately represent the notion of conceptual state
and conceptual state transition,
because it ignores the asymmetric nature of the current state:
we are at state $k_0$,
we are not at state $k_1$
(although we may want to transition there).
The notion of ``current state'' is well represented by extensional linkage.

{\em Extensional linkage\/}
$\lambda_\bullet 
 \type 
 \product{{\cal L}}{{\cal L}} \rightarrow [0,1]$,
which ranges between $0$ and $1$,
is a fuzzy measure of the implication between concepts.
This asymmetric measure of linkage or implication,
which is defined as the ratio of the sizes of extents
\small
\[ \lambda_\bullet(k_0,k_1) 
= \frac{\| \mbox{extent}\left({k_0 \wedge k_1}\right) \|}
{\| \mbox{extent}(k_0) \|}
= \frac{\| \mbox{extent}(k_0) \;\cap\; \mbox{extent}(k_1) \|}
{\| \mbox{extent}(k_0) \|}
= \frac{\sigma_\bullet(k_0,k_1)}
{\| \mbox{extent}(k_0) \|}, \]
\normalsize
measures the implication ``$k_0$ implies $k_1$''.
Extensional linkage can be informally interpreted as a measure of relevance:
$\lambda_\bullet(k_0,{-})$ measures the strength of connection,
transitional strength, or relevance,
from conceptual state $k_0$ to other conceptual 
states.
Extensional linkage can be formally interpreted as 
the probability of $k_1$ conditioned on $k_0$;
that is, the conditional probability $p(k_1 {\mid} k_0)$.
The maximum measure of linkage or implication
represents a strict, full, or Boolean measure of linkage or implication
``$k_0$ strictly implies $k_1$''.
This occurs at the concept lattice order
\begin{center}$\begin{array}{r@{\hspace{4mm}}c@{\hspace{4mm}}l} 
	k_0 \leq k_1
	& \mbox{iff}
	& \lambda_\bullet(k_0,k_1) = 1.
\end{array}$\end{center}
So,
conceptual linkage subsumes 
the hierarchical linkage of the lattice order of concepts.
Extensional linkage $\lambda_\bullet$ 
can be represented by 
a square matrix of real numbers in the interval $[0,1]$,
whose dimension is the cardinality of the set of conceptual classes in the lattice of the conceptual knowledge system.


\section{Conceptual Neighborhood}

The lattice of concepts in a conceptual knowledge system
is intuitively regarded as an environment or conceptual space.
There are two dual senses or modes 
for the idea of 
a ``local neighborhood'' of a concept within its conceptual space.
These two senses of neighborhood are closely bound up with
the two modes of conceptual linkage.
The {\em extensional neighborhood\/} ${\cal N}_\bullet(k)$
of a ``seed'' concept $k$
regards the concept as an attribute:
it fuses the intent of the concept as a collective attribute
and
distributes the extent downward over a local neighborhood lattice.
Precisely defined,
the conceptual knowledge system of the extensional neighborhood
is the restriction of the global conceptual knowledge system
to the extent of the concept
--- all objects not in the extent are ignored.
In terms of conceptual structure,
for any conceptual state $k$
the local extensional neighborhood concept lattice 
${\cal N}_\bullet(k)$
is the restriction of the global lattice ${\cal L}$
by means of the {\em meet restriction\/} operation
${k \wedge ( \mbox{\bf .} )}$.
The meet restriction operation
${k \wedge ( \mbox{\bf .} )}
 \type
 {\cal L} \rightarrow {\cal N}_\bullet(k)$
is right adjoint right inverse to a monotonic map
${\cal N}_\bullet(k) \rightarrow {\cal L}$
which embeds 
the extensional neighborhood lattice into the global lattice.
This means that
meet restriction is meet-preserving since it is right adjoint,
and surjective since it is right inverse.
%

The size of the extensional neighborhood 
depends upon 
the genericity of the seed concept.
The extensional neighborhood of the top concept is very large,
the entire global conceptual knowledge system.
The extensional neighborhood of the bottom concept is very small,
having only one concept. 
Since the extent is usually much smaller than 
the entire set of objects of the global conceptual knowledge system,
the concept neighborhood notion gives a drastic reduction in 
the size of the conceptual space.
The collection of all attributes which label the ``root'' node 
(top concept)
is the intent of the seed concept.
At the opposite pole,
any attribute
which labels the bottom node
is extensionally disjoint from the seed concept in the global lattice
(except for any ``solution objects''
--- objects which satisfy \underline{all} properties).
We can loosely regard the extensional neighborhood lattice line diagram
to be a hierarchy
labeled by the extent of $k$.
These extensional objects are distributed 
over this local neighborhood lattice
by means ``distinguishing attributes''.
By definition,
these attributes are not in the intent of $k$.
This observation forms the basis
for local browsing in the extensional mode via intensional difference.

Between any two concepts $k_0$ and $k_1$ in a concept lattice ${\cal L}$
is the {\em intensional difference\/}
$\partial^\bullet(k_0,k_1) 
 = \mbox{intent}(k_1) \setminus \mbox{intent}(k_0)$,
an asymmetric measure
which records those attributes of $k_1$ that are not attributes of $k_0$.
Elements in $\partial(k_0,k_1)$ are attributes 
which ``distinguish'' $k_1$ from $k_0$.
The intensional difference 
$\partial^\bullet 
	\type \product{{\cal L}}{{{\cal L}^{\rm op}}} \rightarrow \wp{M}
	= \quadruple{\wp{M}}{\supseteq}{\cup}{\emptyset}$
is a generalized metric or distance function,
which satisfies 
the zero law
$\emptyset \supseteq \partial^\bullet(k,k)$
and the triangle law
$\partial^\bullet(k_0,k_1) \cup \partial^\bullet(k_1,k_2)
\supseteq \partial^\bullet(k_0,k_2)$.
All lattice order information is contained in 
the intensional difference, since
\begin{center}$\begin{array}{r@{\hspace{4mm}}c@{\hspace{4mm}}l} 
	k_1 \leq k_0
	& \mbox{iff}
	& \partial^\bullet(k_0,k_1) = \emptyset.
\end{array}$\end{center}
The intensional difference
is 
the basis for the idea of a dictionary definition.
A word (thought of as an object concept)
is defined 
by restricting or specializing
a superordinate (more generic) concept 
by means of a collection of distinguishing properties:
	a concept $k_1$ is-a concept $k_0$ 
	which satisfies	all attributes $m$ 
	in the intensional difference $\partial^\bullet(k_0,k_1)$.
For example,
``a tree is a plant which is woody, perennial and has a main stem.''
Here
``tree'' is the concept being defined (definiendum),
 ``plant'' is the superordinate concept,
and ``woody'', ``perennial'', and ``main stem'' are in the intensional difference.
In the same fashion,
in a conceptual lattice,
we can then think of the collection of differentiating attributes
as representing the difference 
between a defined concept and the superordinate concept.

The {\em intensional difference\/} measure
$\delta^\bullet 
	\type \product{{\cal L}}{{{\cal L}^{\rm op}}} \rightarrow \aleph
	= \quadruple{\aleph}{\geq}{+}{0}$
is also a generalized metric,
which satisfies 
the zero law
$0 \geq \delta^\bullet(k,k)$
and the triangle law
$\delta^\bullet(k_0,k_1) + \delta^\bullet(k_1,k_2)
\geq \delta^\bullet(k_0,k_2)$.
It is a measure of the difference between any two concepts 
according to their intensional difference cardinality.
The intensional difference measure 
is defined by the formulae
\small
\[ \delta^\bullet(k_0,k_1)
= \| \partial^\bullet(k_0,k_1) \|
= \| \mbox{intent}(k_1) \setminus \mbox{intent}(k_0) \| \]
\normalsize
for any two concepts $k_0,k_1$ in a concept lattice ${\cal L}$.
Again,
all lattice order information is contained in 
the intensional difference measure, since
\begin{center}$\begin{array}{r@{\hspace{4mm}}c@{\hspace{4mm}}l} 
	k_1 \leq k_0
	& \mbox{iff}
	& \delta^\bullet(k_0,k_1) = 0.
\end{array}$\end{center}
The minimum measure $0$
occurs when concept $k_1$ is at or below concept $k_0$
in the main lattice.
This occurs when 
no attribute distinguishes concept $k_1$ from concept $k_0$,
although there might be 
an attribute which distinguishes concept $k_0$ from concept $k_1$.
The intensional difference measure
counts the number of distinct distinguishing attributes.
It measures how distinguished $k_1$ is from $k_0$.

A {\em ranked order\/} \mbox{$\pair{{\cal X}}{\rho}$}
consists of
a partially ordered set ${\cal X} = \pair{X}{\leq}$
and an monotonic map 
$\rho \type {\cal X} \rightarrow \aleph = \{0,1,\cdots\}$
to the natural numbers called a {\em ranking\/}.
Ranked orders can be displayed by inverse image
$\rho^{-1}(n) = \{ x \in X \mid \rho(x) = n \}$,
either directly
$\left( \rho^{-1}(0),\rho^{-1}(1),\cdots,\rho^{-1}({\rm max}) \right)$
or in reverse order
$\left( \rho^{-1}({\rm max}),\rho^{-1}({\rm max}{-}1),\cdots,\rho^{-1}(0) \right)$.
Ranked orders are used here as
reduced representations for concept lattices.
They are most useful for 
browsing via the local conceptual neighborhoods,
in 
either the extensional mode
where we browse over the views and attributes of the global lattice,
or the intensional mode
where we browse over the views and objects.
Table~\ref{rankings:Plan1} displays the extensional mode rankings
for the conceptual view ``Plan1''.
The upper panel displays the extensional similarity ranking
at conceptual state ``Plan1'',
a reduced representation for 
the global document conceptual space displayed in Table~\ref{document:cks}.
Here
concepts ``Document'' and ``Object'' have merged in the ranking
with concept ``Plan1'',
whereas the opposite ranking pole shows that concept ``Plan2'' is extensionally disjoint from concept ``Plan1''.
This ranking displays 
all of the irreducible conceptual views and attribute concepts
in the document universe ${\cal D}$.
The lower panel displays the intensional difference ranking
of concept ``Plan1'',
a reduced representation for 
the local document neighborhood of ``Plan1''.
This ranking displays only 
the extent of concept ``Plan1''.

\footnotesize
\begin{table}
\caption{Extensional Mode Rankings for the Conceptual View ``Plan1''}
\label{rankings:Plan1}
\begin{center}
\begin{tabular}{c}
$\mbox{\begin{tabular}{c}{\bf Global}\\{\bf Scope}\end{tabular}}\;\left\{\;
\fbox{\begin{tabular}{|@{\hspace{1mm}}r@{\hspace{2mm}}l|} \hline
	\multicolumn{2}{|c|}{
\begin{tabular}{c}Extensional Similarity Ranking\\
$\sigma_\bullet(\mbox{Plan1},{-})$\end{tabular}
} \\ \hline\hline
	3 &
\mbox{\rule{0mm}{4mm}
$\{[\mbox{Object}],[\mbox{Document}],[\mbox{Plan1},\mbox{project=plan1}]\}$
}
	\\
	2 & 
\mbox{\rule{0mm}{4mm}
$\{\}$
}
	\\
	1 &
\mbox{\rule{0mm}{4mm}
$\{[\mbox{PostScript},\mbox{format=postscript}],[\mbox{format=text}]\}$ }
	\\
	0 &
\mbox{\rule[-2mm]{0mm}{6mm}
$\{[\mbox{Plan2},\mbox{project=plan2}]\}$
}
	\\ \hline
\end{tabular}}
\right.$
\\ \\
$\mbox{\begin{tabular}{c}{\bf Local}\\{\bf Scope}\end{tabular}}\;\left\{\;
\fbox{\begin{tabular}{|@{\hspace{1mm}}r@{\hspace{2mm}}l|} \hline
	\multicolumn{2}{|c|}{
\begin{tabular}{c}Intensional Difference Ranking\\
$\delta^\bullet(\mbox{Plan1},{-})$\end{tabular}
} \\ \hline\hline
	0 &
\mbox{\rule{0mm}{4mm}
$\{[\mbox{Plan1}]\}$
}
	\\
	1 &
\mbox{\rule[-2mm]{0mm}{6mm}
$\{[\mbox{plan1.ps}],[\mbox{notes0.txt}]\}$
}
	\\ \hline
\end{tabular}}
\right.$
\end{tabular}
\end{center}
\end{table}
\normalsize

{\em Conceptual browsing\/} is browsing over conceptual linkage.
It is dual mode (extensional versus the intensional) 
and 
dual scope (global versus local).
Extensional and intensional mode are temporally disjoint,
whereas
global scope is antecedent to local scope both logically and temporally:
choose a mode;
first browse globally in that mode 
and then browse locally in the same mode.
Theoretically,
conceptual browsing ranges over all concepts,
with concepts being represented by internal indexes.
Practically,
conceptual browsing ranges only over named concepts:
objects, attributes, and conceptual views.
In extensional mode we browse over concepts by restriction to their extents.
In intensional mode we do just the lattice dual
--- we browse over concepts by restriction to their intents.
Browsing in the global scope means browsing over the global concept lattice,
whereas
browsing in a local scope means browsing over a local neighborhood concept lattice.
Conceptual browsing is summarized in Table~\ref{modes}.

If a concept lattice is regarded as a form of database structure,
then conceptual browsing can be used for database access,
as in information retrieval \cite{Wi94}.
In this approach conceptual linkage is used for processing queries.
A query in intensional mode involves only 
the attributes of the formal context under consideration.
By definition,
an {\em intensional query\/} is a subset of attributes.
It can be identified with a new temporary ``goal query'' object
which has been added to the formal context.
The goal query object is regarded to be 
the current conceptual state
for browsing in intensional mode.
Then,
intensional linkage ranking is 
a vector of similarity coefficients,
each coefficient measuring 
the closeness of a concept to the goal query.
Either conceptual views and objects can be display as a ranking,
or those conceptual views and objects can be returned
whose similarity coefficient is above a given threshold.
By duality,
an {\em extensional query\/} is a subset of objects.
The query is identified with a new temporary ``goal query'' attribute,
which is regarded to be 
the current conceptual state
for browsing in extensional mode.
The objects in the query are regarded as prototypes.
Extensional queries correspond to 
a prototype representation for categories (conceptual classes).
Issuing an extensional query
results in returning similarity measures
between conceptual classes 
and the collective prototype of the query's objects.

\renewcommand{\thempfootnote}{\fnsymbol{mpfootnote}}
\begin{table}
\caption{The Process of Conceptual Neighborhood Browsing}
\label{modes}
\begin{center}
\fbox{
\begin{minipage}{11.8cm}
\hspace{2mm}
\begin{tabular}[t]{c}
\mbox{ } \\
\fbox{\begin{tabular}{|c|}\hline
{\bf Extensional Mode} \\
Current Concept:
$k$ \\ \hline\hline
\hspace{1mm}
\begin{minipage}{10.2cm}
\vspace{4mm}
\begin{description}
	\item[Global Scope:]
		Display the extensional similarity ranking
		\[ \sigma_\bullet(k,{-}) 
		= \| \mbox{extent}(k) \cap \mbox{extent}({-}) \| \]
		for global lattice ${\cal L}$.
		Selection of a concept from this display
		to be the next conceptual state
		will continue extensional mode browsing.
	\item[Local Scope:]
		Display the intensional difference ranking
		\[ \delta^\bullet(k,{-}) 
		= \mbox{intent}({-}) \setminus \mbox{intent}(k) \]
		for local lattice ${\cal N}_\bullet(k)$.
		Selection of a concept from this display
		to be the next conceptual state
		will switch to intensional mode browsing.
\end{description}
\vspace{2mm}
\footnotetext{In this mode
$k$ is either a conceptual view or an attribute concept.}
\end{minipage}
\hspace{1mm}
\\ \hline
\end{tabular}}
\\
\begin{tabular}{c@{\hspace{2.5cm}}c}
\begin{picture}(40,40)
\thicklines
\put(18,8){\line(0,1){20}}
\put(22,8){\line(0,1){20}}
\put(20,32){\line(1,-2){5}}
\put(20,32){\line(-1,-2){5}}
\thinlines
\end{picture}
&
\begin{picture}(40,40)
\thicklines
\put(18,12){\line(0,1){20}}
\put(22,12){\line(0,1){20}}
\put(20,8){\line(1,2){5}}
\put(20,8){\line(-1,2){5}}
\thinlines
\end{picture}
\end{tabular}
\\
\fbox{\begin{tabular}{|c|}\hline
{\bf Intensional Mode} \\
Current Concept: $k$ \\ \hline\hline
\hspace{1mm}
\begin{minipage}{10.2cm}
\vspace{4mm}
\begin{description}
	\item[Global Scope:]
		Display the intensional similarity ranking
		\[ \sigma^\bullet(k,{-}) 
		= \| \mbox{intent}(k) \cap \mbox{intent}({-}) \| \]
		for global lattice ${\cal L}$.
		Selection of a concept from this display
		to be the next conceptual state
		will continue intensional mode browsing.
	\item[Local Scope:]
		Display the extensional difference ranking
		\[ \delta_\bullet(k,{-}) 
		= \mbox{extent}({-}) \setminus \mbox{extent}(k) \]
		for local lattice ${\cal N}^\bullet(k)$.
		Selection of a concept from this display
		to be the next conceptual state
		will switch to extensional mode browsing.
\end{description}
\vspace{1pt}
\footnotetext{In this mode 
$k$ is either a conceptual view or an object concept.}
\end{minipage}
\hspace{1mm}
\\ \hline
\end{tabular}}
\\ \mbox{ }
\end{tabular}
\footnotetext{At any time during browsing, 
we can request
either the definition of a conceptual view,
the definition of an attribute,
or the summary information for an object.
In either the extensional or intensional mode
we can issue a corresponding query.}
\end{minipage}}
\end{center}
\end{table}


\section{Conceptualization Processes}

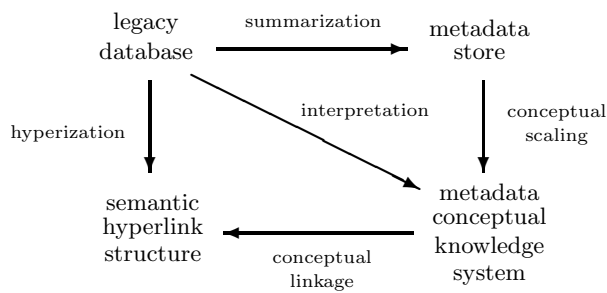
\begin{figure}
\caption{Conceptual Interpretation and Database Hyperization}
\label{database}
\begin{center}
\begin{picture}(200,120)(-41,-30)
\setlength{\unitlength}{1.2pt}
\put(-24,50){\shortstack{legacy\\database}}
\put(80,50){\shortstack{metadata\\store}}
\put(-23,-14){\shortstack{semantic\\hyperlink\\structure}}
\put(81,-20){\shortstack{metadata\\conceptual\\knowledge\\system}}
\put(-52,25){\mbox{\scriptsize{hyperization}\normalsize}}
\put(22,60){\mbox{\scriptsize{summarization}\normalsize}}
\put(28,-22){
\shortstack{
\mbox{\scriptsize{conceptual}\normalsize}
\\
\mbox{\scriptsize{linkage}\normalsize}
}
}
\put(37,32){
\mbox{\scriptsize{interpretation}\normalsize}
}
\put(102,24){
\shortstack{
\mbox{\scriptsize{conceptual}\normalsize}
\\
\mbox{\scriptsize{scaling}\normalsize}
}
}
\thicklines
\put(-8,43){\vector(0,-1){29}}
\put(5,45){\vector(2,-1){72}}
\put(75,-5){\vector(-1,0){60}}
\put(13,53){\vector(1,0){60}}
\put(97,43){\vector(0,-1){29}}
\thinlines
\end{picture}
\end{center}
\end{figure}

The intuitive idea behind hypertext is ``semantic connection'' \cite{tbl92}.
Currently,
hyperlink creation is done manually at document creation time \cite{Wi94}.
There are two problems with this manual approach:
\begin{itemize}
	\item The document creator (writer, publisher) 
		may inadvertantly omit 
		some important and meaningful semantic connections.
	\item Legacy data (pre-HTML documents)
		needs enormous manual effort
		in order to convert to hypertextual form.
\end{itemize}
Figure~\ref{database} gives a high-level description 
of processes involved in 
the conceptual organization and representation of 
the information in legacy databases.
The interpretation process \cite{Ke94b} is a composite of 
summarization followed by conceptual scaling
\cite{GaWi89,KeNe94},
	$\left( \mbox{interpretation} \right)
	=
	\left( \mbox{summarization} \right)
	\circ
	\left( \mbox{conceptual scaling} \right)$.
Summarization is the abstraction and construction of metadata objects
from actual data.
The gathering component of the Harvest system \cite{BDHMS94}
is a good example of summarization.
Conceptual scaling,
also called relational data filtration,
is a user-oriented process 
for customizing and building a faceted representation of information
based upon user interest profiles, etc.
Here a user may refer to
either a single individual,
a small group of individuals,
or even a whole community.
Conceptual scaling uses type-structured standing queries,
known as conceptual scales \cite{GaWi89},
alerts, continuous queries,
or SDI (selective dissemination of information) \cite{Ro87}.

The hyperization process is a process of automatic web archiving.
As such,
it answers the concerns expressed above about manual web creation.
Actually,
hyperization could represent 
either the batch process of web archiving
or the interactive process of web guidance during client browsing.
Hyperization is a composite of 
interpretation followed by conceptual linkage,
	$\left( \mbox{hyperization} \right)
	=
	\left( \mbox{interpretation} \right)
	\circ
	\left( \mbox{conceptual linkage} \right)$.
As depicted in Figure~\ref{database},
conceptual linkage involves 
the automatic creation of {\em crisp\/} web hyperlink structure
by a reduction process of crispification. 
There is,
however,
information loss in just the creation of crisp web hyperlinks.
In this sense,
it is better to remain at 
the higher level of the conceptual knowledge system,
rather than reducing to web hyperlink structure.
At the higher level of the conceptual knowledge system,
conceptual linkage richly expresses 
conceptual structure and semantic content.

\begin{figure}
\caption{Conceptual Structuring of Hypertext Incidence}
\label{web}
\begin{center}
\begin{picture}(190,40)(-30,-13)
\setlength{\unitlength}{1.2pt}
\put(-14,-6){\shortstack{web\\hyperlink\\structure}}
\put(83,-13){\shortstack{hyperlink\\conceptual\\knowledge\\system}}
\put(34,14){\mbox{\scriptsize{generation}\normalsize}}
\put(40,-11){\mbox{\scriptsize{readout}\normalsize}}
\thicklines
\put(28,7){\vector(1,0){45}}
\put(73,-2){\vector(-1,0){45}}
\thinlines
\end{picture}
\end{center}
\end{figure}
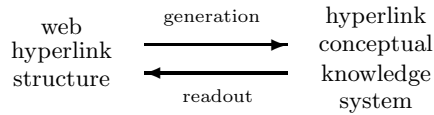

Figure~\ref{web} describes the equivalence 
between the hyperlink structure of the Web
and its representation as a conceptual knowledge system.
In the application of the conceptual knowledge system model 
to Web hyperlinkage,
both objects and attributes are Web objects (HTML documents, images, etc.).
There are two dual interpretations for hyperlink incidence:
(cross-referential)
one Web object has a second Web object as an attribute
when
the first points to the second;
and
(hierarchical, such as gopher-space)
the opposite incidence \cite{tbl92,Wi94}.
The web-cks equivalence in Figure~\ref{web}
is mediated through the inverse passages
of concept generation and incidence readout:
	$\left( \mbox{generation} \right) 
		\circ \left( \mbox{readout} \right)
	\equiv 
	\left( \mbox{identity} \right)$
and
	$\left( \mbox{readout} \right) 
		\circ \left( \mbox{generation} \right)
	\equiv
	\left( \mbox{identity} \right)$.
These inverse passages comprise the standard process diagram from formal concept analysis,
here applied to the incidence relationships of Web hyperlinks.
The process of concept generation results in 
a conceptual representation for Web hyperlinkage.

\begin{figure}
\caption{Enriched Web Construction}
\label{combined}
\begin{center}
\begin{picture}(270,170)(-65,-35)
\setlength{\unitlength}{1.2pt}
\put(-46,63){
$\mbox{web} 
\left\{ 
\begin{array}{c} 
\mbox{hyperlinks} \\ 
{+} \\ 
\mbox{data}
\end{array}
\right.$
}
\put(110,75){\shortstack{hyperlink\\conceptual\\knowledge\\system}}
\put(64,28){\shortstack{metadata\\conceptual\\knowledge\\system}}
\put(-17,-15){\shortstack{enriched\\web\\structure}}
\put(107,-22){\shortstack{enriched\\conceptual\\knowledge\\system}}
\put(-46,25){\mbox{\scriptsize{hyperization}\normalsize}}
\put(36,86){\mbox{\scriptsize{generation}\normalsize}}
\put(20,45){\mbox{\scriptsize{generation}\normalsize}}
\put(43,-22){
\shortstack{
\mbox{\scriptsize{conceptual}\normalsize}
\\
\mbox{\scriptsize{linkage}\normalsize}
}
}
\put(131,34){\mbox{\scriptsize{apposition}\normalsize}}
\put(125,35){\circle*{4}}
\thicklines
\put(20,70){\vector(4,1){82}}
\put(20,62){\vector(4,-1){35}}
\put(-3,46){\vector(0,-1){35}}
\put(125,35){\line(-3,1){12}}
\put(125,35){\line(1,6){5}}
\put(125,35){\vector(0,-1){20}}
\put(97,-5){\vector(-1,0){76}}
\thinlines
\end{picture}
\end{center}
\end{figure}
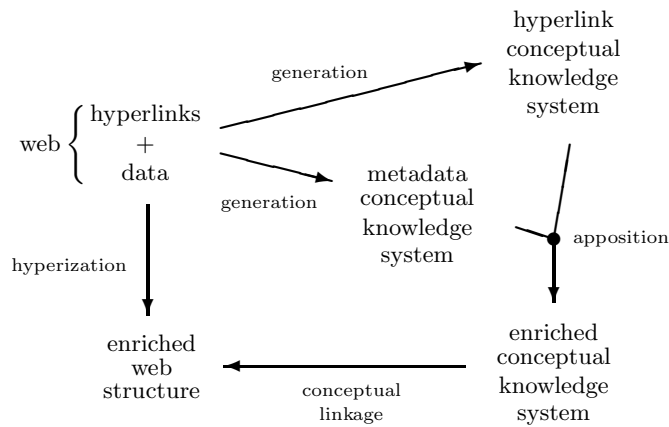

The information in the Web can be split into two distinct aspects:
hyperlinkage and document content \cite{tbl92,Wi94}.
Distinct processes applied to these two aspects
generate distinct conceptual representatons.
The semantic constraint between hyperlinkage and document content
can be applied later during 
the process of conceptual scaling and data filtration.
Substitution of the non-linked content of a web 
for the legacy database component in Figure~\ref{database} 
describes a process for the conceptual representation of Web content.
In Figure~\ref{combined} we describe an enriched process
which combines 
the hyperlink conceptual representation of Figure~\ref{web}
with the metadata conceptual representation of Figure~\ref{database}.
This enriched combining process
uses a standard combinator from formal concept analysis called apposition.
At the incidence matrix level of a conceptual knowledge system,
apposition is a summing process,
whereas at the concept lattice level it is a constrained producting process.
The same comments that we made above 
about crisp conceptual linking 
are true here also:
it is better to do conceptual linkage 
at the enriched conceptual knowledge system level
--- here there is no loss of information
and a richer conceptual expression.

\section{Summary, Implementation, and Future Work}

This chapter has discussed 
two approaches for making use of automatic classification techniques:
web archive construction and user navigational guidance (conceptual browsing).
Automatic classification provides the foundation
for the automatic generation of local web hypertextual structure
based upon summarized and conceptually scaled information about objects.
Manual specification of conceptual connectivity,
such as for ordinary hyperlinks and conceptual views,
can automatically be incorporated.
Automatic classification also provides the foundation
for guidance and analysis 
during user browsing and concept link traversal 
via the World Wide Web
over any community's information space.
In summary,
formal concept analysis is a principled foundation 
for classification, organization, and indexing in NIDR systems.
By using ideas from formal concept analysis,
Web hyperlinks can be elaborated into Web conceptual links,
and Web hyperspace can be coherently organized as Web conceptual space.

By the time of publication,
many of the ideas discussed in this chapter
will have been implemented
in the NIDR system called WAVE which was mentioned above.
These ideas include
formal contexts,
concept lattices,
conceptual knowledge systems,
conceptual optimization,
conceptual linkage,
and
conceptual browsing.
Development of the WAVE system
will provide
a preliminary answer to the research question:
``What is the appropriate architecture for a digital library?''
It will be demonstrated
in the distributed context of the World Wide Web,
by using 
both the technique of automatic classification 
and 
the notion of a conceptual knowledge system,
that the WAVE system
provides the kernel architecture for a digital library.
A critical measure of success for the WAVE system
will be	the ability to {\em understand} a user's intentions.
The understanding of intentions is a very deep research question, 
and as Dennis Reinhardt has pointed out to the first author
(private communication), 
machine understanding will not surpass human capability in this area 
during the course of this research. 
However, 
the WAVE system could augment human understanding 
with the ability to {\em express} a user's intentions.
This sense of ``understanding'' a user's intentions 
will be a critical factor in the success of the WAVE system. 
Other measures, 
such as how customizable, how adaptable, or 
how flexible the system is for the user, 
are subordinate strategies 
which will aid the ability to express the user's intentions. 

Both on-going and future work can be discussed in terms of 
three processes for 
the conceptualization of networked information resources,
as diagrammed in Figure~\ref{database}:
summarization, conceptual scaling, and conceptual linkage.
The first process,
summarization,
has been implemented as
the front component of a NIDR system,
where meta-information is extracted.
An important example of a summarization processor 
is the gatherer component of the Harvest system.
The third process,
conceptual linkage,
is now being implemented as
the first phase (funded) 
of the WAVE system development.
This phase,
called WAVE{\it Guide\/},
will replace the broker indexing component of the Harvest system,
extending broker capabilities
by adding dynamic and customizable knowledge organization techniques.
WAVE{\it Guide\/} will be used for 
interactive information analysis and browsing guidance
during exploratory search by client Web browsers
over a community's information space.
The second process,
conceptual scaling,
will next year be implemented as
the second phase 
of WAVE system development.
This phase,
called WAVE{\it Form\/},
will represent the process of faceted analysis
which occurs in library science classification.
It also corresponds to the design of user interest profiles
in current awareness services \cite{Ro87}.

%
%

%

\begin{thebibliography}{55}
%
\bibitem{Wi82}
Wille, R.:
Restructuring Lattice Theory: An Approach Based on Hierarchies of Concepts.
Ordered Sets,
I.\ Rival (ed.),
Reidel,
Dordrecht-Boston
(1982)
445--470
%
\bibitem {Ro87}
Rowley, J.:
Organising Knowledge: An Introduction to Information Retrieval.
Gower,
Aldershot, Hants, England
(1987)
%
\bibitem {GaWi89}
Ganter, B., Wille, R.:
Conceptual Scaling.
Applications of Combinatorics and Graph Theor
y in the Biological and Social Sciences,
Springer,
New York
(1989),
F.\ Roberts (ed.),
139--167
%
\bibitem{Wi92}
Wille, R.:
Concept Lattices and Conceptual Knowledge Systems.
Computers and Mathematics with Applications,
vol.\ 23,
493--522
(1992)
%
\bibitem{WyTa92}
Wynar, B., Taylor, A.:
Introduction to Cataloging and Classification, 8th ed.
Libraries Unlimited,
Englewood, Colorado
(1992)
%
\bibitem {tbl92}
Berners-{L}ee, T., Cailliau R., Groff J., Pollerman, B.:
{W}orld-{W}ide {W}eb: The Information Universe.
CERN,
Geneva, Swizerland,
(1992)
%
\bibitem{Wi94}
Wilson, E.:
Hypertext Libraries:
The Automatic Production of Hypertext Documents.
Research in Humanities Computing,
S.\ Hockey and N.\ Ide (eds.),
232-246 (1994)
%
\bibitem{DeEm94}
Deutsch, P., Emtage, A.:
Publishing Information on the Internet with {A}nonymous {FTP}.
Bunyip Information Systems Inc., 
(May 1994)
%
\bibitem{BDBCP94}
Bowman, M., Dharap, C., Baruah, M., Camargo, B., Potti, S.:
A File System for Information Management.
Proceedings of the Conference on Intelligent Information Management Systems,
(June 1994)
%
\bibitem{BDHMS94}
Bowman, M., Danzig, P., Hardy, D., Manber, U., Schwartz, M.:
Harvest: A Scalable, Customizable Discovery and Access System.
technical report CU-CS-732-94,
University of Colorado,
(July 1994)
%
\bibitem{Me94}
Mealling, M.:
Encoding and Use of Uniform Resource Characteristics.
Internet Engineering Task Force ({IETF}),
Internet draft document
draft-ietf-uri-urc-spec-00.txt,
(July 1994)
%
\bibitem {KeNe94}
Kent, R.E., Neuss, C.:
Creating a 3{D} {W}eb {A}nalysis and {V}isualization {E}nvironment.
Electronic Proceedings of 
the Second International {W}orld {W}ide {W}eb Conference ({WWWF}'94), {M}osaic and the {W}eb,
(October 1994)
%
\bibitem{BDHMS94}
Bowman, M., Danzig, P., Hardy, D., Manber, U., Schwartz, M.:
The Harvest Information Discovery and Access System.
Electronic Proceedings of 
the Second International {W}orld {W}ide {W}eb Conference ({WWWF}'94), {M}osaic and the {W}eb,
(October 1994)
%
\bibitem{Ke94b}
Kent, R.E.:
Enriched Interpretation.
Proceedings of the Third International Workshop 
on Rough Sets and Soft Computing ({RSSC}'94),
(November 1994)
%
\bibitem {NeKe95}
Neuss, C., Kent, R.E.:
Conceptual Analysis of Resource Meta-information.
Electronic Proceedings of 
the Third International {W}orld {W}ide {W}eb Conference ({WWW}'95),
(April 1995)
%
\bibitem {KeBo95}
Kent, R.E., Bowman, M.:
Digital Libraries, Conceptual Knowledge Systems, and the Nebula Interface.
technical report, 
Transarc Corporation,
Pittsburgh
(April 1995)
submitted for publication
\end{thebibliography}
\end{document}